\documentclass{article}

\usepackage{arxiv}

\usepackage[utf8]{inputenc} 
\usepackage[T1]{fontenc}    
\usepackage{hyperref}       
\usepackage{url}            
\usepackage{booktabs}       
\usepackage{amsfonts}       
\usepackage{amsmath}
\usepackage{nicefrac}       
\usepackage{microtype}      
\usepackage{cleveref}       
\usepackage{lipsum}         
\usepackage{graphicx}
\usepackage[numbers]{natbib}
\usepackage{doi}
\usepackage{parskip}

\usepackage{algorithm}
\usepackage{algpseudocode}

\DeclareMathOperator*{\argmax}{arg\,max}

\title{ACHO: Adaptive Conformal Hyperparameter Optimization}

\date{}

\author{Riccardo Doyle \\ London, United Kingdom \\
 \texttt{r.doyle.edu@gmail.com}
}

\renewcommand{\headeright}{}
\renewcommand{\undertitle}{}

\hypersetup{
pdftitle={ACHO Adaptive Conformal Hyperparameter Optimization - Doyle},
pdfsubject={cs.LG},
pdfauthor={Riccardo Doyle},
pdfkeywords={hyperparameter optimization, conformal prediction, automl, deep learning},
}

\begin{document}
\maketitle

\begin{abstract}
    Several novel frameworks for hyperparameter search have emerged in the last decade, but most rely on strict, often normal, distributional assumptions, limiting search model flexibility. This paper proposes a novel optimization framework based on upper confidence bound sampling of conformal confidence intervals, whose weaker assumption of exchangeability enables greater choice of search model architectures. Several such architectures were explored and benchmarked on hyperparameter search of random forests and convolutional neural networks, displaying satisfactory interval coverage and superior tuning performance to random search.
\end{abstract}


\section{Introduction} \label{Introduction}

Identifying optimal model parameters is deeply desirable for high prediction performance in machine learning, but challenging due to non-convexity and expensive search costs.
Common approaches involving grid search - exhaustive iterative search of a confined parameter interval - or random search \citep{JMLR:v13:bergstra12a} - random sampling from a broader parameter space – display complimentary weaknesses and form no expectation of hyperparameter performance ahead of search. Focus has instead centered on search frameworks capable of forming hyperperameter performance expectations prior to sampling, generally dominated by Sequential Model-Based Optimization (SMBO) \citep{SMBO}. Early applications \citep{SMBO, NIPS2011_86e8f7ab} resulted in positive outperformance on expert consensus across a range of benchmarked datasets, leveraging Gaussian Process or Tree-structured Parzen estimators. Further expansions of the framework included search cost inclusion as an optimization criterion \citep{NIPS2012_05311655}, early forms of online resource allocation and distributed search \citep{NIPS2013_f33ba15e}, unwanted parameter space pruning \citep{10.1007/978-3-319-23525-7_7}, or replacement of single estimators with ensemble methods \citep{DBLP:journals/corr/LacosteLLM14}. Though alternative approaches \citep{JMLR:v18:16-558} have been proposed and popularized, SMBO-based search remains one of the most widespread, non-naïve search methods in the training of complex machine learning predictors, with frequent applications in competitions \citep{pmlr-v133-turner21a} and package releases \citep{pmlr-v28-bergstra13, optuna_2019, JMLR:v21:18-223}. Its normally distributed architecture allows for a range of robust, probability based acquisition functions, but limits point estimators’ functional forms to generally weaker -- and in the case of Gaussian Processes, slower training -- learners.

In this study we modify the classic SMBO structure to make use of less constraining, conformal interval based acquisition functions with no distributional assumptions beyond exchangeability of point estimator outputs. The latter will be further relaxed by means of adaptive intervals to allow for covariate shift. The proposed approach retains the benefits of a fitted predictor and sequential framework, but offers wider flexibility in the choice of estimator functional forms, allowing for more complex -- and expectantly better fitting -- model architectures than conventionally used Gaussian Processes.

\section{Conformal Prediction Review}\label{CP Overview}

Conformal prediction \citep{JMLR:v9:shafer08a} is a distribution free framework for obtaining valid prediction intervals on exchangeable data. In a split conformal context \citep{lei-split-cp}, taking some training set \(X_{train} ,Y_{train} =\{(X_i,Y_i) \mid {i\in \mathcal{I}_{train}, \mathcal{I}_{train} \not\subset \mathcal{I}_{val}}\}\) and validation set \(X_{val} ,Y_{val} =\{(X_i,Y_i) \mid {i\in \mathcal{I}_{val}, \mathcal{I}_{val} \not\subset \mathcal{I}_{train}}\}\), a conformal interval for some new prediction of some estimator \(\hat Y(X)\) fitted on the training set can be obtained as:
\begin{equation}
I(X)=[\hat Y(X)\pm q_{1-\alpha}(D_{val})]
\end {equation}
Where \(q_{1-\alpha}\) is the \(1-\alpha\) quantile of \(D_{val}\) and \(D_{val}\) is a multiset of exchangeable non-conformity scores. The latter quantify the divergence between an estimator's outputs and its target on a held-out validation set. In regression tasks it is common to set this to the estimator's absolute validation set residuals \(\{ |Y_{i}-\hat Y(X_{i})| \mid i \in 
\mathcal{I}_{val} \}\).

The \(1-\alpha\) parameter in earlier quantile notation corresponds to the desired coverage level of the conformal interval, controlling its width. For a given \(1-\alpha\) however, interval width is invariant to \(X\), rendering it unsuitable as a measure of uncertainty to guide sampling in a hyperparameter search framework.

\subsection{Locally Weighted Conformal Prediction (LWCP)}\label{LWCP Overview}

To produce intervals that vary with the covariate space, the previous section's absolute deviations can be weighted \citep{Lei1} by the outputs of some conditional uncertainty estimator \(U(X)\), producing the following updated non-conformity score formulation:
\begin{equation}
D_{val}=\{ |Y_{i}-\hat Y(X_{i})| / U(X_{i}) \mid i \in \mathcal{I}_{val}\}
\end{equation}
And the following updated conformal interval function:
\begin{equation}
I(X)=[\hat Y(X)\pm U(X)q_{1-\alpha}(D_{val})]
\end{equation}
Where higher uncertainty at a given \(X\) results in a larger interval and vice versa. The uncertainty measure predicted by \(U(X)\) may take several desired forms, including conditional residual spread or conditional spread in \(Y\).

\subsection{Conformalized Quantile Regression (CQR)}\label{CQR Overview}

Locally weighted approaches produce valid, but symmetrical conformal intervals, which may overstate the interval size required to obtain the same coverage where deviations are either primarily positive or primarily negative relative to the point estimator line. This is particularly relevant to hyperparameter search, where a high concentration of deviation below the point estimation is detrimental, but would increase sampling odds.

To remedy this, we can replace the previous sections' point and uncertainty estimators with quantile regression \citep{10.2307/1913643}, directly estimating conditional quantile bounds during training.
The latter involves replacing a conventional regression model's mean squared error loss with pinball loss \(L_\beta\):
\begin{equation}
    L_\beta(u_i) =
    \begin{cases}
    u_i \beta & \text{if \(u_i > 0\)} \\
    u_i(\beta-1) & \text{if \(u_i \leq 0\)}
    \end{cases}
\end{equation}
Where \(u_i\) is the absolute error between model-predicted and observed outputs and \(\beta\) is the desired conditional quantile value to be predicted, with a theoretical perfect quantile regression estimator \(\hat Q_{\beta}(X)\) guaranteeing \(P(Y \leq \hat Q_{\beta}(X=x)) \mid X=x) = \beta\).

A prediction interval for some new observation \(X\) at some desired coverage level \(1-\alpha\) could then be trivially generated by fitting two quantile regression estimators for some pre-specified symmetrical lower and upper quantile levels \(\alpha/2\) and \(1-(\alpha/2)\), with the predictions of each regression forming the bounds of the interval:
\begin{equation}
I(X) = [\hat Q_{\alpha/2}(X), \hat Q_{1-(\alpha/2)}(X)]
\end{equation}
This approach however does not provide coverage guarantees on unseen data. To add calibration, the interval generating process can be extended using conformalization \citep{NEURIPS2019_5103c358}.
Assuming the two aforementioned quantile estimators were fitted on some training data \(X_{train}, Y_{train}\), one can produce non-conformity scores for their combined interval on validation data \(X_{val} ,Y_{val}\) according to:
\begin{equation}
D_{val} = \{max(\hat Q_{\alpha/2}(X_{i})-Y_{i}, Y_{i}- \hat Q_{1-(\alpha/2)}(X_{i})) \mid i \in \mathcal{I}_{val}\}
\end{equation}
A conformal prediction interval can then be obtained by adjusting the initial quantile estimates by the \(1-\alpha\) and \(\alpha\) quantiles of the multiset of validation deviations \(D_{val}\):
\begin{equation}
I(X) = [\hat Q_{\alpha/2}(X) - q_{1-\alpha}(D_{val}), \hat Q_{1-(\alpha/2)}(X) + q_{1-\alpha}(D_{val})]
\end{equation}
This study employs quantile loss adapted versions of Gradient Boosted Machines and the non-quantile loss, non-parametric method of Quantile Regression Forests \citep{JMLR:v7:meinshausen06a} to guide hyperparameter search.

\subsection{Adaptive Conformal Prediction}\label{Adaptive CP Overview}
All previously outlined conformal prediction frameworks require exchangeability of non-conformity scores.
In sequential hyperparameter optimization, a hyperparameter \(\theta_{t+1}\) is sampled for evaluation if a conditional hyperparameter performance estimator \(\hat{\phi}(\theta)\) trained on all previously sampled hyperparameters and their performances, assigns it the highest expected performance.
Non-conformity scores generated by this process are thus not exchangeable, as, while the relationship between true performance and hyperparameter choice \(P(\phi|\theta)\) is unchanged across sampling episodes, the greedy sampling distribution \(P(\theta)\) changes after each, causing non-exchangeable \(\phi\) realizations by way of covariate shift. 

To retain validity, intervals can be adjusted adaptively between sampling episodes to account for distributional shift \citep{gibbs2021adaptive}. Formally, rather than it being static, we update the miss-coverage level \(\alpha\) after each successive observation \(Y_{t+1}\) based on:
\begin{equation}
\alpha_{t+1} = \alpha_t + \gamma(\alpha - \epsilon_t)
\end{equation}
Where \(\gamma\) is some custom learning rate and \(\epsilon_t\) is some misscoverage indicator for the successive observation \(\phi_{t+1}\) on the conformal prediction interval constructed for it, defined as:

\begin{equation}
\epsilon_t = 
\left\{ 
  \begin{array}{ c l }
    1      & \quad \textrm{if } \phi_{t+1} \notin I_t(\theta_{t+1}) \\
    0      & \quad \textrm{otherwise}
  \end{array}
\right.
\end{equation}

The framework has the effect of solving for a validation quantile that yields the originally desired quantile level on the now shifted out of sample set (based on the observed extent of the shift in \(\phi\) alone, not \(\theta\)). Note in a split conformal setting, if the distributional shift is too large, there may not be a quantile that satisfies this equivalence.

\section{Adaptive Conformal Hyperparameter Optimization (ACHO)}\label{ACHO Methodology}

Let \(X_i, Y_i\) be the data available to tune a given machine learning architecture. Let us further sub-set this into training and validation partitions:
\begin{equation}
X_{train} ,Y_{train} =\{(X_{i},Y_{i}) \mid {i\in \mathcal{I}_{train}, \mathcal{I}_{train} \not\subset \mathcal{I}_{val}}\}
\end{equation}
\begin{equation}
X_{val} ,Y_{val} =\{(X_{i},Y_{i}) \mid {i\in \mathcal{I}_{val}, \mathcal{I}_{val} \not\subset \mathcal{I}_{train}}\}
\end{equation}

\subsection{Preliminary Random Search}\label{ACHO - RS}
Prior to any conformal methodology, a random search framework samples \(n\) hyperparameter configurations \(\{\theta_t\}_{t=1}^n\) without replacement from some finite configuration set \(C\), where \(|C| > n\) . In turn, each configuration is used to fit a different estimator \(\hat Y_t(X)\) on training set \(X_{train},Y_{train}\), with a resulting validation performance per configuration \(\theta_t\) of:
\begin{equation}
\phi_t=L(Y_{val},\hat Y_t(X_{val}))
\end{equation}
Where \(L\) is some evaluation metric this study sets to either mean squared error or accuracy, depending on the nature of the data. To simplify optimization, in remaining methodology we take \(L\) to return larger values for better performance, as would be the case when using accuracy to evaluate the model's outputs. Following initial random search, a set of hyperparameter to performance pairs \(\{(\theta_t, \phi_t)\}_{t=1}^n\) is obtained. The pairs constitute a sample of initial observations on which conformal estimators can be fitted to guide successive search. The number of samples required is arbitrary, with this study setting it to either \(t=20\) or \(t=25\). Subsequent methodology differs based on whether a locally weighted or quantile conformal framework is adopted, with each being covered in a separate section.

\subsection{Locally Weighted Conformal Inference (LWCI)}\label{ACHO - LWCP}
For a locally weighted conformal search framework, let us split the hyperparameter to performance pairs \(\{(\theta_t, \phi_t)\}_{t=1}^n\) into training and validation sets:
\begin{equation}
    \theta_{train} ,\phi_{train} =\{(\theta_t,\phi_t) \mid {t\in \mathcal{T}_{train}, \mathcal{T}_{train} \not\subset \mathcal{T}_{val}}\}
\end{equation}
\begin{equation}
    \theta_{val} ,\phi_{val} =\{(\theta_t,\phi_t) \mid {t\in \mathcal{T}_{val}, \mathcal{T}_{val} \not\subset \mathcal{T}_{train}}\}
\end{equation}
And further split the training set into two additional sub-sets:
\begin{equation}
    \theta_{train'} ,\phi_{train'} =\{(\theta_t,\phi_t) \mid {t\in \mathcal{T}_{train'}, \mathcal{T}_{train'} \in \mathcal{T}_{train}, \mathcal{T}_{train'} \not\subset \mathcal{T}_{val'}}\}
\end{equation}
\begin{equation}
    \theta_{val'} ,\phi_{val'} =\{(\theta_t,\phi_t) \mid {t\in \mathcal{T}_{val'}, \mathcal{T}_{val'} \in \mathcal{T}_{train}, \mathcal{T}_{val'} \not\subset \mathcal{T}_{train'}}\}
\end{equation}
Next, we fit some model on the sub-training data \(\theta_{train'}, \phi_{train'}\) to produce an estimator of performance conditional on a chosen configuration, denoted as \(\hat \phi(\theta)\). Sampling subsequent configurations based on this estimator alone would result in greedy sub-optimal exploration of the parameter space. To capture uncertainty around its predictions, locally weighted conformal intervals can be generated according to section \ref{LWCP Overview}. 
Taking the conditional mean absolute deviation of the estimator's residuals (denoted hereafter as \(V\)) as our measure of uncertainty, we can fit a second model on the sub-validation set pairs of configurations and residual deviations \(\theta_{val'} ,V_{val'}\), yielding a conditional uncertainty estimator \(\hat V(\theta)\). Note the estimator is fitted on the sub-validation set, as \(V\) observations must be generated from \(\hat \phi(\theta)\)'s residuals and using residuals from data \(\hat \phi(\theta)\) previously used for training would produce bias.

Having fitted both a point and uncertainty estimator, a multiset of validation set non-conformity scores can then be generated according to:
\begin{equation}
    D_{val}=\{ |\phi_{t}-\hat \phi(\theta_{t})| / \hat V(\theta_{t}) \mid t \in \mathcal{T}_{val}\}
\end{equation}
Consequently, we can define a conditional prediction interval for any of \(\hat \phi(\theta)\)'s predictions for a specified coverage level \(1-\alpha\) as:
\begin{equation}
I(\theta) = [\hat \phi(\theta) \pm \hat V(\theta)q_{1-\alpha}(D_{val})]
\end{equation}
This interval enhances the predictions of the standalone greedy point estimator \(\hat \phi(\theta)\) to include variable uncertainty bounds, conditional on the configuration space.

\subsection{Conformalized Quantile Inference (CQI)}\label{ACHO - CQR}
For a conformalized quantile search framework, referring to configuration and performance subsets from equations 13 and 14, let us fit two quantile regression models on the training data \(\theta_{train}, \phi_{train}\) to predict lower and upper performance bounds conditional on the configuration space. As outlined in section \ref{CQR Overview}, for a desired coverage level \(1-\alpha\), the interval must be generated by a lower bound estimator \(\hat Q_{\alpha/2}(\theta)\) and a higher bound estimator \(\hat Q_{1-(\alpha/2)}(\theta)\), producing the following multiset of validation non-conformity scores:
\begin{equation}
D_{val} = \{ max(\hat Q_{\alpha/2}(\theta_{t})-\phi_{t}, \phi_{t}-\hat Q_{1-(\alpha/2)}(\theta_{t})) \mid t \in \mathcal{T}_{val}\}
\end{equation}
Which results in a conformalized quantile regression-derived search interval for some candidate configuration \(\theta\) of:
\begin{equation}
I(\theta) = [\hat Q_{\alpha/2}(\theta) - q_{1-\alpha}(D_{val}), \hat Q_{1-(\alpha/2)}(\theta) + q_{1-\alpha}(D_{val})]
\end{equation}

\begin{algorithm}[hb]
	\caption{ACHO under a CQI framework} 
	\begin{algorithmic}[0] \label{ACHO pseudo code}

        \Require \(\alpha \in [0,1], \gamma > 0, m > n\)
	    \State \(X_{train}, X_{val}, Y_{train}, Y_{val}\longleftarrow\ train\_val\_split(X, Y)\)
            \State \(C = \{\theta_i\}_{i=1}^m\)
            \State Randomly sample \(H \subset C, |H| = n\)
            \State Take \(P\) to contain the paired performances \(\phi\) of each \(\theta\) in \(H\)
	    \For {$t=n+1:m$}
            \State \(H_{train},\ H_{val},\ P_{train},\ P_{val}\longleftarrow train\_val\_split(H,\ P)\)
            \State Fit \(\hat Q_{\alpha/2}(\theta)\) on \(H_{train}, P_{train}\)
            \State Fit \(\hat Q_{1 - (\alpha/2)}(\theta)\) on \(H_{train}, P_{train}\)
            \State \(D_{val} = \{ max(\hat Q_{\alpha/2}(\theta_j)-\phi_j, \phi_j-\hat Q_{1-(\alpha/2)}(\theta_j)) \mid \theta_j \in H_{val}, \phi_j \in P_{val} \} \)
            \State \(\theta_{t+1}= \argmax_\theta (\{\hat Q_{1-(\alpha/2)}(\theta) + q_{1-\alpha}(D_{val}) \mid \theta \in C, \theta \notin H\})\)
            \State Fit \(\hat Y(X)\) on \(X_{train}, Y_{train}\) with maximal hyperparameter configuration \(\theta_{t+1}\)
            \State \(\phi_{t+1}=L(\hat Y(X_{val}), Y_{val})\)
            \If {$\hat Q_{(1-\alpha)/2}(H_{val}) < \phi_{t+1} < \hat Q_{\alpha + (1-\alpha)/2}(P_{val}) $}
                \State \(\epsilon_t = 0\) 
            \Else
                \State \(\epsilon_t = 1\) 
            \EndIf
            \State \(\alpha_{t} = \alpha_{t-1} + \gamma(\alpha - \epsilon_t)\)

            \State \(P \leftarrow P \cup (\phi_{t+1})\)
            \State \(H \leftarrow H \cup (\theta_{t+1})\)

            \If {$\phi_{t+1} > \phi_{best}$}
                \State Update best search performance \(\phi_{best} \leftarrow \phi_{t+1}\)
                \State Update best hyperparameter configuration \(\theta_{best} \leftarrow \theta_{t+1}\)
            \EndIf
		\EndFor
	\end{algorithmic} 
\end{algorithm}

\subsection{Conformal Search}\label{ACHO - Search}

By producing an expected performance interval for each unsampled configuration \(\{\theta \mid \theta \in C, \theta \notin \{\theta_t\}_{t=1}^n \}\) through \(I(\theta)\), the next best configuration to search can be selected through Upper Confidence Bound (UCB) sampling \citep{auer2002finite} of the interval, formalized as:
\begin{equation}
\theta_{t+1}= \argmax_\theta (\{\hat \phi(\theta) + \hat V(\theta)q_{1-\alpha}(D_{val}) \mid \theta \in C, \theta \notin \{\theta_t\}_{t=1}^n \})
\end{equation}
When producing intervals via a locally weighted framework, or:
\begin{equation}
\theta_{t+1}= \argmax_\theta (\{\hat Q_{1-(\alpha/2)}(\theta) + q_{1-\alpha}(D_{val}) \mid \theta \in C, \theta \notin \{\theta_t\}_{t=1}^n\})
\end{equation}
When producing intervals via a conformalized quantile framework.

A new estimator \(\hat Y_{t+1}(X)\) is then fitted on this configuration and its resulting configuration and loss pair \((\theta_{t+1}, \phi_{t+1})\) is appended to the original \(\{(\theta_t, \phi_t)\}_{t=1}^n\) set.  Steps in either sections \ref{ACHO - RS} and \ref{ACHO - LWCP} or \ref{ACHO - RS} and \ref{ACHO - CQR} are then repeated on the new expanded configuration and loss set, constituting a sequential hyperparameter configuration search framework. At each successive refitting of the conformal estimators in sections \ref{ACHO - LWCP} or \ref{ACHO - CQR}, the coverage level \(1-\alpha\) used to construct the interval function \(I(\theta)\) is updated according to:
\begin{equation}
\alpha_{t+1} = \alpha_t + \gamma(\alpha - \epsilon_t)
\end{equation}
Accounting for any covariate shift arising from the transition between random search and epsilon-greedy sampling, or subsequent smaller distributional shifts thereafter.
A pseudo-code summary of the above methodology using a conformalized quantile approach can be found in Algorithm 1.

\section{Benchmarking and Results}\label{Results}

\subsection{Random Forest Tuning}\label{Results - RF}

Performance is first tested on the parameters of a Random Forest base model with a searchable hyperparameter space comprised of 1000 randomly generated combinations of individual parameter values reported in Table \ref{RF hyperparams}. 

\begin{table}[hb]
\caption{Individual hyperparameter values comprising searchable hyperparameter space of base Random Forest model.}\label{RF hyperparams}%
\centering
\begin{tabular}{@{}ll@{}}
\toprule
Hyperparameter & Search Value\\
\midrule
Number of Estimators & [10, 20 ..., 90, 100, 150, 200, 300, 400] \\
Split Minimum Samples (\% of Samples) & [0.005, 0.01, 0.05, 0.1, 0.2, 0.3] \\
Leaf Minimum Samples (\% of Samples) & [0.005, 0.01, 0.05, 0.1, 0.2, 0.3] \\ 
Maximum Features (\% of Features) & [0.1, 0.2, .. 1] \\

\bottomrule
\end{tabular}
\end{table}

We consider the following real-life benchmark dataset:  
\begin{itemize}
    \item HOUSING: Regression dataset with house price target variable, 8 census based features and 20,640 observations based on the 1990 California census \citep{RePEc:eee:stapro:v:33:y:1997:i:3:p:291-297}. Data is obtained via \textit{scikit-learn} \citep{scikit-learn}.
\end{itemize}
And the following synthetic datasets:
\begin{itemize}
    \item FRIEDMAN-1*, FRIEDMAN-2*, FRIEDMAN-3*: Artificial regression datasets with uniformly distributed, some times redundant, independent features. Target variables for each are derived from varying deterministic transformations of the features followed by an additive noise term. The original composition of each is described in \citep{21a96583-326f-3b42-a89a-6c4389d8dfd0}, though minor modifications to the total number of observations and noise terms are made, with the asterisk suffix in this study's naming marking the distinction. Data is obtained via \textit{scikit-learn} \citep{scikit-learn} with construction and parameters detailed in Appendix \ref{Appendix - Synthetics}.
    \item HYPERCUBE: Artificial classification dataset with balanced binary labels and clustered features centered at the vertices of a hypercube. The original generating algorithm is described in \citep{Guyon2003DesignOE} and data is obtained from \textit{scikit-learn} \citep{scikit-learn} with parameters detailed in Appendix \ref{Appendix - Synthetics}.
\end{itemize}

\begin{figure}[t]
\centering
\includegraphics[scale=0.27]{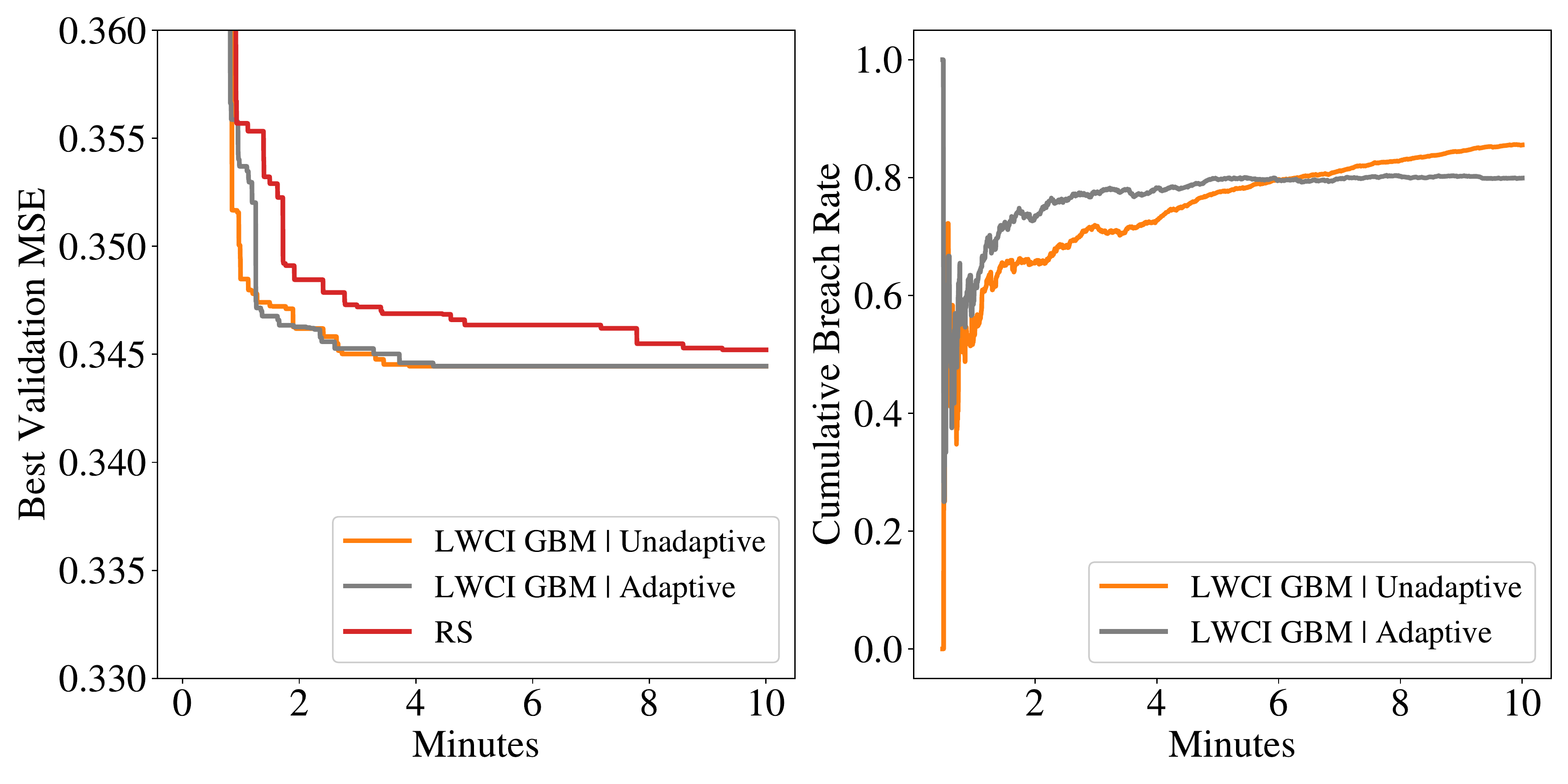}
\caption{\textbf{Left}: Best tuning mean squared error (MSE) achieved over search time on HOUSING data through either a locally weighted (LWCI) gradient boosted machine (GBM) framework at coverage \(1-\alpha=20\%\), or random search (RS). LWCI GBM search is repeated once with unadaptive intervals and once with adaptive intervals. Values are centisecond averages of 10 randomly seeded runs of each framework. \textbf{Right}: Cumulative conformal interval breach rate over search time on LWCI GBM sampling intervals.}\label{adaptive plot}
\end{figure}

Optimization performance of the ACHO algorithm across a variety of parameters is explored on each dataset. 

Fig. \ref{adaptive plot} reports HOUSING data validation mean squared error (MSE) for a locally weighted conformal inference (LWCI) framework with a gradient boosted machine (GBM) architecture used in fitting both \(\hat \phi(\theta)\) and \(\hat V(\theta)\) estimators outlined in section \ref{ACHO - LWCP}. The framework is run with both adaptive and unadaptive intervals at coverage \(1-\alpha=20\%\). Performance is reported over search time and accompanied by a random search (RS) counterpart. 
Both adaptive and unadaptive LWCI GBM frameworks outperform RS for the duration of search, with a final validation MSE of 0.344 and 0.345 respectively (Appendix \ref{Appendix - RF Results}). Impact of adaptive intervals on coverage is highly corrective, with breach rates between sampled observations and their pre-sampling intervals converging to their expected level (\(\alpha=80\%\)) with initially less undershooting, and subsequently no overshooting, in the adaptive variant.

\begin{figure}[b]
\centering
\includegraphics[scale=0.27]{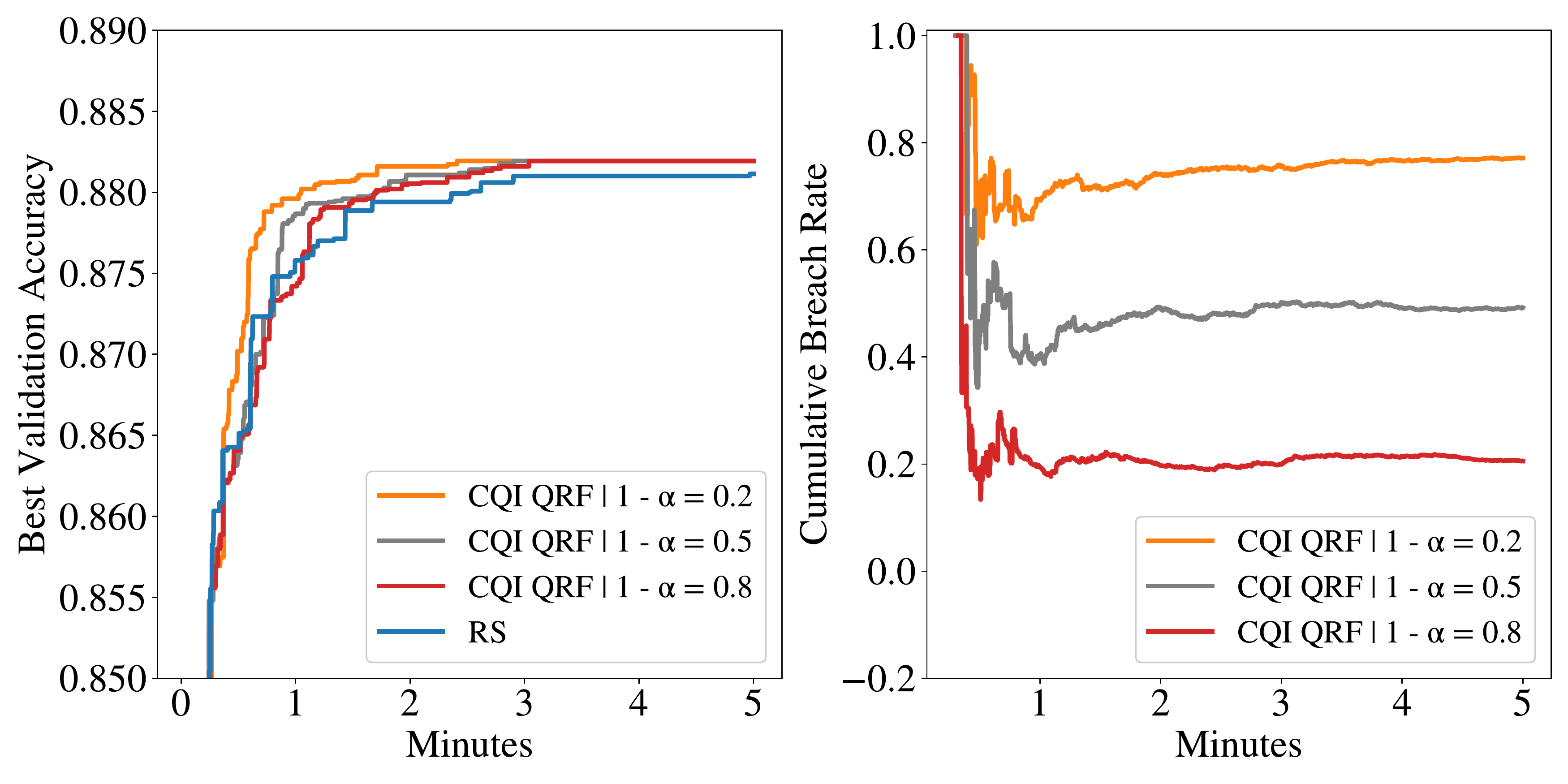}
\caption{\textbf{Left}: Best tuning accuracy achieved over search time on HYPERCUBE data through either a conformalized quantile (CQI) framework with a quantile regression forest (QRF) estimator or random search (RS). QRF search is repeated at various coverage levels \(1-\alpha\). Values are centisecond averages of 10 randomly seeded runs of each framework. \textbf{Right}: Cumulative conformal interval breach rate over search time on QRF sampling intervals.}\label{alpha plot}
\end{figure}

Fig. \ref{alpha plot} reports HYPERCUBE data validation accuracy for a conformalized quantile (CQI) framework with a quantile regression forest (QRF) estimator at different coverage levels \(1-\alpha\). 
A larger coverage indicates a preference for exploration over exploitation during upper confidence bound sampling (UCB) via increased sampling interval width. It is thus of interest to analyse how different trade offs perform. We note, though all CQI QRF variants achieve the same, better than random, final validation accuracy of 88.19\% (Appendix \ref{Appendix - RF Results}) in Fig. \ref{alpha plot}, there are clear differences in search path between each, with the most exploitative 20\% coverage framework outperforming random search the most, and the earliest, followed by worsening convergence speeds as coverage grows. This relationship may be reversed for a dataset with more complex loss surfaces, where exploration may be better rewarded. We continue to note that realized coverage on the conformal search intervals is performant, with near exact end of run breach rates for the 80\% coverage variant (20.59\%, Appendix \ref{Appendix - RF Results}) and 50\% coverage variant (49.20\%, Appendix \ref{Appendix - RF Results}), and satisfactory ones for the 20\% variant (77.12\%, Appendix \ref{Appendix - RF Results}).

\begin{figure}[t]
\centering
\includegraphics[scale=0.39]{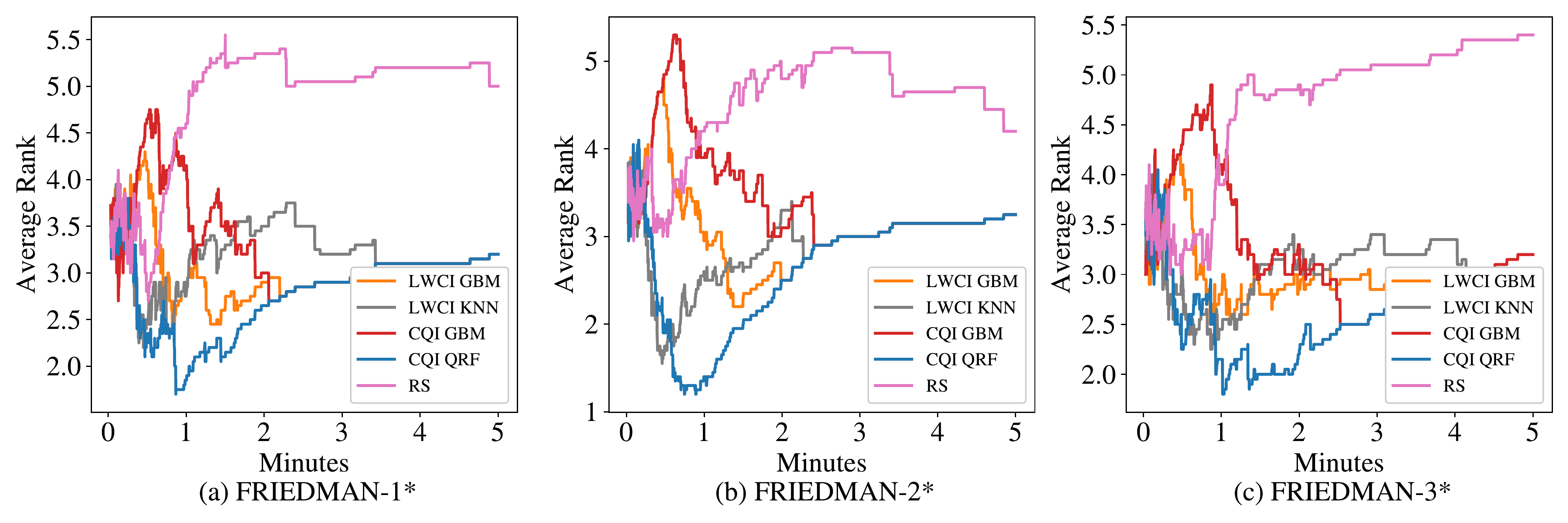}
\caption{Average rank of ACHO frameworks alongside random search (RS) over search time on the (a) FRIEDMAN-1* (b) FRIEDMAN-2* (c) FRIEDMAN-3* synthetic regression datasets at \(1-\alpha=20\%\) coverage. Values are centisecond averages of 10 randomly seeded runs of each framework.}\label{rank plot}
\end{figure}

A sole choice of conformalized (CQI) QRF and locally weighted (LWCI) GBM frameworks have been explored in preceding parameter-specific benchmarks. To provide greater visibility on the impact of estimator architecture choice in ACHO frameworks, Fig. \ref{rank plot} reports the average performance rank of five such variants over search time across the FRIEDMAN-1*, FRIEDMAN-2* and FRIEDMAN-3* datasets. We note regardless of estimator framework, random search underperforms for the majority of search time. 
Prior to elaborating on more granular patterns, it is worth noting this paper's implementation of ACHO enforces each search estimator undergoes a mandatory tuning run upon first training. This run can reoccur with reduced parameters or at irregular frequencies depending on its run-time impact, but always occurs once with fixed parameters upon obtaining the first \(t\) random samples in ACHO. Under this framework, slower training estimators will incur a greater run-time delay compared to faster ones in the seconds following the end of the ACHO random sampling period, affecting their rank regardless of estimator quality.
Considering this, we note conformalized quantile regression forest (CQI QRF) frameworks outperform their counterparts across datasets. This is likely due to a combination of strong estimation performance and reduced training or tuning delay, as QRF's training time is scale invariant to the number of quantiles to estimate. LWCI and CQI GBM frameworks perform positively due to GBM's high estimation quality, but experience tuning delays, as LWCI frameworks involve both point and variance estimator training, and CQI GBM runtime scales with quantile count.
LWCI KNN matches early QRF search performance, but loses rank as search progresses. KNN's training speed and more limited parameter space likely enables its early outperformance, but its lower estimation quality results in limited gains past its runtime benefit.

Several other estimator architectures may be tested (including mixed architectures for locally weighted frameworks) and best architecture patterns seen above may not necessarily carry forward to different types of datasets; particularly where scale is a factor, as estimator tuning delays tend to relative insignificance as base dataset size increases. A more thorough grid benchmarking of various dataset types may provide useful priors for choice of architecture in real life application.

\subsection{Convolutional Neural Network Tuning}\label{Results - CNN}

We further test ACHO performance on a convolutional neural network (CNN) to validate usage on a more complex architecture. The searchable space is defined by 1000 randomly generated combinations of individual parameter values reported in Table \ref{CNN hyperparams}.

\begin{table}[h]
\caption{Individual hyperparameter values comprising searchable hyperparameter space of base convolutional neural network model.}\label{CNN hyperparams}%
\centering
\begin{tabular}{@{}ll@{}}
\toprule
Hyperparameter & Search Value\\
\midrule
Solver & [Adam, SGD] \\ 
Learning Rate & [0.0001, 0.0005, 0.001, 0.005, 0.01, 0.05, 0.1] \\ 
Drop Out Rate & [0.1, 0.2, .. 0.9] \\ 
Number of Convolutions in Given Layer & [16, 32, .. 64] \\
Number of Layers & [2, 3] \\
First Dense Layer:
Number of Neurons & [100, 200, 512]\\
Second Dense Layer:
Number of Neurons & [0, 50, 100]\\
\bottomrule
\end{tabular}
\end{table}

We consider the following image recognition datasets for benchmarking:
\begin{itemize}
    \item MNIST: Handwritten digit recognition dataset with 10 classes, 60,000 training examples and 10,000 validation examples \citep{726791}.
    \item FASHION-MNIST: Black and white object recognition dataset with 10 clothing item classes, 60,000 training examples and 10,000 validation examples \citep{Xiao2017FashionMNISTAN}.
\end{itemize}

\begin{figure}[h]
\centering
\includegraphics[scale=0.39]{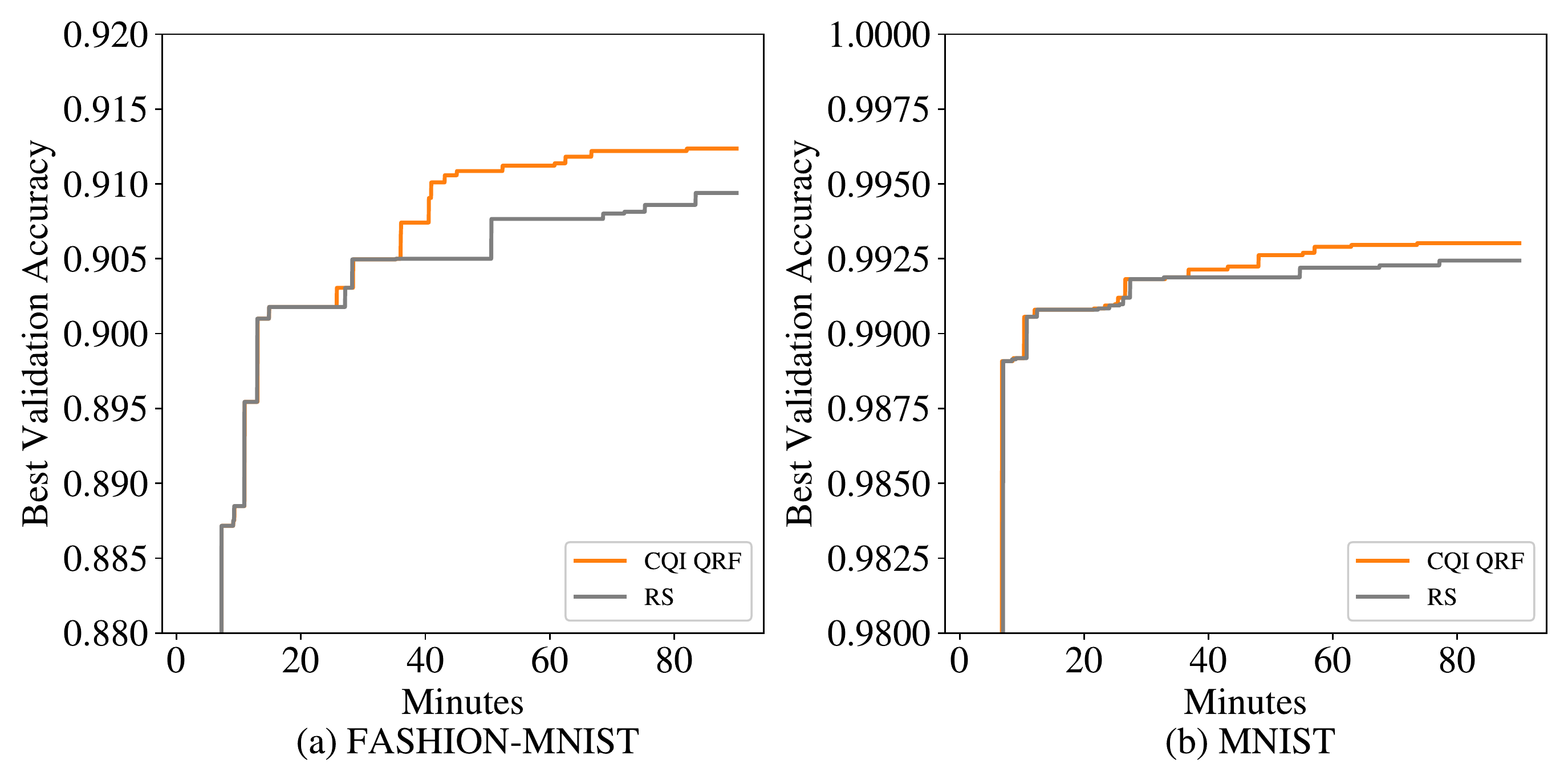}
\caption{Best tuning accuracy achieved over search time on MNIST and FASHION-MNIST data by either a conformalized (CQI) quantile regression forest (QRF) framework at coverage \(1-\alpha=20\%\) or random search (RS). All values are centisecond averages of 5 randomly seeded runs of each framework.}\label{cnn plot}
\end{figure}

Fig. \ref{cnn plot} summarizes search validation accuracy achieved on the MNIST and FASHION-MNIST datasets by a conformalized quantile (CQI) framework with a quantile regression forest (QRF) estimator at \(1-\alpha=20\%\) coverage. Performance
is reported over search time and accompanied by random search (RS). Following its shared preliminary random sampling period, we note QRF outperforms RS on both datasets, with a final validation accuracy of 91.24\% versus 90.94\% (Appendix \ref{Appendix - CNN Results}) on the FASHION-MNIST dataset, and 99.30\% versus 99.24\% (Appendix \ref{Appendix - CNN Results}) on the MNIST dataset. Though not plotted, conformal uncertainty estimation remains satisfactory, with end-of-run validation set breach rates of 72.99\% on the FASHION-MNIST dataset and 72.06\% (Appendix \ref{Appendix - CNN Results}) on the MNIST dataset. Realized breaches exhibit greater deviations than reported on tabular benchmarks, likely due to the significantly lower number of sampling events on the more computationally expensive convolutional benchmarks.

Strong ACHO outperformance is thus detectable on both simple tabular and more complex image-based datasets.

\section{Conclusion}

This study introduced a novel optimization framework for hyperparameter selection based on conformal prediction. Performance across a range of benchmarked datasets spanning tabular classification, tabular regression and image recognition was meaningfully superior to random search in both final performance and time to achievement. Empirical coverage of conformal intervals was satisfactory, with most displaying modest or negligible divergences from their theoretical bounds at medium search time horizons.
Further performance improvements could be achieved with the introduction of early stopping logic based on expected improvement inference, the inclusion of expected search cost in acquisition function design and the replacement of simple adaptive conformal intervals with variations that do not suffer from quantile bound limitations where covariate shift is excessive.

\newpage
\begin{appendix}

\section{Benchmark Results}\label{Appendix - Results}

\begin{table}[h]
\caption{Performance of ACHO frameworks and random search (RS) in tuning a Random Forest architecture across range of specified datasets. Each row's accuracy, MSE (Mean Squared Error) or breach rate (on conformal intervals) is an average of 10 randomly seeded runs of the row's framework.}
\label{Appendix - RF Results}%
\footnotesize
\centering
\begin{tabular}{p{2.2cm}p{2.3cm}p{1.4cm}p{1.3cm}p{1.5cm}p{1.5cm}p{1.5cm}}
\toprule
Dataset & Search \newline Framework & Target \newline Coverage \newline (\(1-\alpha\)) & Adaptive & Final \newline Validation \newline MSE & Final \newline Validation \newline Accuracy & Final \newline Breach \newline Rate\\

\midrule 
HYPERCUBE & CQI QRF & 20\% & True &  ~ & 88.19\% & 77.12\% \\
HYPERCUBE & CQI QRF & 50\% & True &  ~ & 88.19\% & 49.20\% \\
HYPERCUBE & CQI QRF & 80\% & True &  ~ & 88.19\% & 20.59\% \\
HYPERCUBE & RS &  ~  &  ~  &  ~ & 88.11\% &  ~  \\
 
\midrule 
HOUSING & LWCI GBM & 20\% & False & 0.344 &  ~ & 85.53\% \\
HOUSING & LWCI GBM & 20\% & True & 0.344 &  ~ & 79.9\% \\
HOUSING & RS &  ~  &  ~  & 0.345 &  ~ &  ~  \\
 
\midrule 
FRIEDMAN-1* & LWCI GBM & 20\% & True & 3.827 &  ~ & 80.03\% \\
FRIEDMAN-1* & LWCI KNN & 20\% & True & 3.827 &  ~ & 82.71\% \\
FRIEDMAN-1* & CQI GBM & 20\% & True & 3.827 &  ~ & 73.27\% \\
FRIEDMAN-1* & CQI QRF & 20\% & True & 3.827 &  ~ & 77.86\% \\
FRIEDMAN-1* & RS &  ~  &  ~  & 3.852 &  ~ &  ~  \\
 
\midrule 
FRIEDMAN-2* & LWCI GBM & 20\% & True & 388.8 &  ~ & 79.38\% \\
FRIEDMAN-2* & LWCI KNN & 20\% & True & 388.8 &  ~ & 81.96\% \\
FRIEDMAN-2* & CQI GBM & 20\% & True & 388.8 &  ~ & 74.24\% \\
FRIEDMAN-2* & CQI QRF & 20\% & True & 388.8 &  ~ & 77.16\% \\
FRIEDMAN-2* & RS &  ~  &  ~  & 391.8 &  ~ &  ~  \\
 
\midrule 
FRIEDMAN-3* & LWCI GBM & 20\% & True & 1.044 &  ~ & 79.68\% \\
FRIEDMAN-3* & LWCI KNN & 20\% & True & 1.044 &  ~ & 81.66\% \\
FRIEDMAN-3* & CQI GBM & 20\% & True & 1.044 &  ~ & 77.03\% \\
FRIEDMAN-3* & CQI QRF & 20\% & True & 1.044 &  ~ & 78.30\% \\
FRIEDMAN-3* & RS &  ~  &  ~  & 1.045 &  ~ &  ~  \\
\bottomrule
\end{tabular}

\end{table}

\begin{table}[h]
\caption{Performance of ACHO frameworks and random search (RS) in tuning a Convolutional Neural Network architecture across range of specified datasets. Each row's accuracy or breach rate (on conformal intervals) is an average of 5 randomly seeded runs of the row's framework.}
\label{Appendix - CNN Results}%
\footnotesize
\centering
\begin{tabular}{p{2.7cm}p{2cm}p{1.5cm}p{1.3cm}p{1.5cm}p{1.5cm}}
\toprule
Dataset & Search \newline Framework & Target \newline Coverage \newline (\(1-\alpha\)) & Adaptive & Final \newline Validation \newline Accuracy & Final \newline Breach \newline Rate\\

\midrule
FASHION-MNIST & CQI QRF & 20\% & True & 91.24\% & 72.99\% \\
FASHION-MNIST & RS &  ~  &  ~  & 90.94\% &  ~  \\
 
\midrule 
MNIST & CQI QRF & 20\% & True & 99.30\% & 72.06\% \\
MNIST & RS &  ~  &  ~  & 99.24\% &  ~  \\

\bottomrule
\end{tabular}
\end{table}

\newpage
\section{Synthetic Dataset Construction}\label{Appendix - Synthetics}

Benchmarking made use of four synthetic datasets. The construction details of the FRIEDMAN group of datasets is summarized below -- with reference to the original construction details in \citep{21a96583-326f-3b42-a89a-6c4389d8dfd0} and more explicitly in \citep{friedman-supporting}, though changes to the total number of observations and noise terms were made:
\begin{itemize}
\item FRIEDMAN-1*: 10,000 observations (which is larger than the original construction and simulates tuning on a medium sized regression dataset) and 10 independent features uniformly distributed along \([0,1]\). Only five of the features are relevant for prediction, with target variable \(Y\) generated according to \(Y = 10 \sin(\pi(X_1 X_2)) + 20 (X_3 - 0.5)^2 + 10 X_4 + 5 X_5 + \epsilon\), where \(\epsilon \sim \mathcal{N}(0, 1)\).

\item FRIEDMAN-2*: 10,000 observations (which is larger than the original construction and simulates tuning on a medium sized regression dataset) and 4 independent features uniformly distributed along the following intervals:
    \begin{equation}
    0 \geq X_1 \leq 100
    \nonumber
    \end{equation}
    \begin{equation}
    20 \geq \frac{X_2}{2 \pi} \leq 280
    \nonumber
    \end{equation}
    \begin{equation}
    0 \geq X_3 \leq 1
    \nonumber
    \end{equation}
    \begin{equation}
    1 \geq X_4 \leq 11
    \nonumber
    \end{equation}
Target variable \(Y\) is generated according to:
\begin{equation}
Y =  (X_1^2 + (X_2 X_3 - (1/ X_2 X_4))^2)^{1/2} + \epsilon
\nonumber
\end{equation}
Where \(\epsilon \sim \mathcal{N}(0, 1)\), which differs from original construction to apply varying signal to noise ratios between datasets.

\item FRIEDMAN-3*: 10,000 observations (which is larger than the original construction and simulates tuning on a medium sized regression dataset) and 4 independent features uniformly distributed along the intervals described in FRIEDMAN-2*, with target \(Y\) generated according to:
\begin{equation}
Y =  \tan^{-1} \frac{(X_2 X_3 - (1/ X_2 X_4))}{X_1} + \epsilon
\nonumber
\end{equation}
Where \(\epsilon \sim \mathcal{N}(0, 1)\), which differs from original construction to apply varying signal to noise ratios between datasets.

\end{itemize}
FRIEDMAN and HYPERCUBE datasets used in Section \ref{Results - RF} were generated via \textit{scikit-learn==1.3.0} \citep{scikit-learn}. The following functions and overrides were used to obtain each:
\begin{itemize}
    \item FRIEDMAN-1*: \textit{sklearn.datasets.make\_friedman1} function with default parameters, except for \textit{n\_samples=10000}, \textit{noise=1}, \textit{random\_state=1234}.
    \item FRIEDMAN-2*: \textit{sklearn.datasets.make\_friedman2} function with \textit{n\_samples=10000}, \textit{noise=1}, \textit{random\_state=1234}.
    \item FRIEDMAN-3: \textit{sklearn.datasets.make\_friedman3} function with \textit{n\_samples=10000}, \textit{noise=1}, \textit{random\_state=1234}.
    \item HYPERCUBE: \textit{sklearn.datasets.make\_classification} function with default parameters, except for \textit{n\_features=10}, \textit{n\_redundant=5}, \textit{n\_informative=5}, \textit{class\_sep=5}, \textit{random\_state=1234}.
\end{itemize}

\end{appendix}
\newpage

\bibliographystyle{unsrt}
\bibliography{main} 

\end{document}